\documentclass[draftcls]{IEEEtran}
\ifCLASSINFOpdf
\else
\fi

\usepackage{flushend}
\usepackage{graphicx}
\usepackage{enumerate,color}
\usepackage[justification=centering]{caption}

\begin{document}

\title{Agnostic Lane Detection}

\author{HOU-Yuenan,~\IEEEmembership{Department of Information Engineering}
}

\markboth{ First year report}
{Liew : Procedure and Guidelines for First-Year Written and Oral Exam}

\maketitle

\IEEEpeerreviewmaketitle

\begin{abstract}
Lane detection is an important yet challenging task in autonomous driving, which is affected by many factors, e.g., light conditions, occlusions caused by other vehicles, irrelevant markings on the road and the inherent long and thin property of lanes. Conventional methods typically treat lane detection as a semantic segmentation task, which assigns a class label to each pixel of the image. This formulation heavily depends on the assumption that the number of lanes is pre-defined and fixed and no lane changing occurs, which does not always hold. To make the lane detection model applicable to an arbitrary number of lanes and lane changing scenarios, we adopt an instance segmentation approach, which first differentiates lanes and background and then classify each lane pixel into each lane instance. Besides, a multi-task learning paradigm is utilized to better exploit the structural information and the feature pyramid architecture is used to detect extremely thin lanes. Three popular lane detection benchmarks, i.e., TuSimple, CULane and BDD100K, are used to validate the effectiveness of our proposed algorithm.
\end{abstract}

\section{Introduction}

Lane detection ~\cite{bertozzi1998gold} plays a pivotal role in autonomous driving because lanes could serve as significant cues for constraining the maneuver of vehicles on roads. However, lane detection is challenging since it is affected by many factors, e.g., light conditions, occlusions caused by other vehicles, the existence of irrelevant markings on the road and the inherent long and thin property of lanes. 

Conventional methods~\cite{borkar2012novel, deusch2012random} usually utilize hand-crafted features to extract lane segments and can perform quite well in the highway driving scenarios. However, these approaches need a good selection of features and have poor generalization ability. Therefore, they cannot be applied to scenarios with varying light conditions and road types. The emergence of deep learning has brought new insights into the task and Convolutional Neural Network (CNN) based methods begin to gain popularity ~\cite{lee2017vpgnet, pan2017spatial, ghafoorian2018gan, chen2017rbnet, hou2018learning}. The inherent and automatic feature extracting ability of CNN eases the complex feature selection process and partially solves the generalization problems. However, the CNN-based methods perform sub-optimally in urban roads where the lane markings are ambiguous or the lanes are severely occluded. Several schemes have been proposed to handle lane detection in urban roads, e.g., performing message passing to better exploit structural information ~\cite{pan2017spatial} or utilizing vanishing points to guide the lane detection task ~\cite{lee2017vpgnet}. These methods can work to some extent but cannot fully solve the problem as they ignore the inherent relationship between the different entities in the driving scenarios. For instance, the areas within two neighbouring lanes (i.e., drivable areas and alternative areas~\cite{yu2018bdd100k}) can serve as a strong indicator for the existence, shape and position of lanes. Besides, these models tend to fail when encountering an arbitrary number of lanes or lane changing since they model lane detection as the semantic segmentation task and each lane is assigned a pre-defined class. Failing to achieve real-time performance is also a drawback of these approaches ~\cite{pan2017spatial, lee2017vpgnet}.  

Therefore, in this study, we propose to use a multi-task learning paradigm to better utilize the structural and contextual information of the driving scenarios. More specifically, besides the traditional lane detection branch, we also borrow the rich structural information from the drivable area detection task and the lane point regression task. The feature pyramid architecture ~\cite{lin2017feature} is also incorporated in our model to handle challenges of detecting extremely thin lanes. To fulfill the real-time requirement, we adopt the light-weight and efficient network, i.e.,  ENet~\cite{paszke2016enet}, as our backbone. To detect conditions with unfixed number of lanes, we follow ~\cite{neven2018towards} and divide the lane detection task into two sub-tasks. The first one is to generate the binary segmentation map which only differentiates the lanes and the background. The second sub-task is to classify the lane pixels into different lane instances (i.e., treat each lane as an instance). Three popular benchmarks, i.e., TuSimple~\cite{tusimple}, CULane~\cite{pan2017spatial} and BDD100K~\cite{yu2018bdd100k}, are selected to validate the effectiveness of our proposed algorithm. Since it is an on-going project, we only report preliminary experimental results on TuSimple and CULane.

\section{Related Work}

Lane detection is conventionally handled via using specialized and hand-crafted features to obtain lane segments. These segments are further grouped to get the final results ~\cite{borkar2012novel, deusch2012random}. These methods are intuitive but have many shortcomings, e.g., requiring complex feature selection process, being lack of robustness and only applicable to relatively easy driving scenarios. 

Recently, deep learning methods ~\cite{lee2017vpgnet, pan2017spatial, ghafoorian2018gan, chen2017rbnet} have been proposed to ease the selection of hand-crafted features as well as greatly improve the models' generalization ability. These approaches usually adopt the dense prediction formulation, i.e., treat lane detection as a semantic segmentation task, where each pixel in an image is assigned with a label to indicate whether it belongs to a lane or not. For example, Pan et al ~\cite{pan2017spatial} propose SCNN, which combines spatial cues with CNN, to generate multi-channel probability maps where the number of channels equals to the number of lanes. However, these methods can only handle scenarios where the number of lanes is pre-defined and fixed, and they often fail when the vehicle is changing lanes. Another drawback is that these approaches could not achieve real-time performance, which impedes them from being used in the real world. 

To overcome these shortcomings, we follow \cite{neven2018towards} and model lane detection as an instance segmentation task. More specifically, the lane detection task is divided into two sub-tasks. The first sub-task is generating a binary segmentation map which differentiates lanes and the background. The second sub-task is classifying each lane pixel into a lane instance. The light-weight network, i.e., ENet ~\cite{paszke2016enet} is used as our backbone to achieve real-time performance. What's more, to utilize the structural and contextual information, we adopt a multi-task learning paradigm in which drivable area detection and lane point regression are incorporated into the original lane detection model. Moreover, the feature pyramid architecture is utilized to detect extremely thin lanes.

\section{Methodology}

\begin{figure*}[t]
  \centering
  \includegraphics[width=1.0\linewidth]{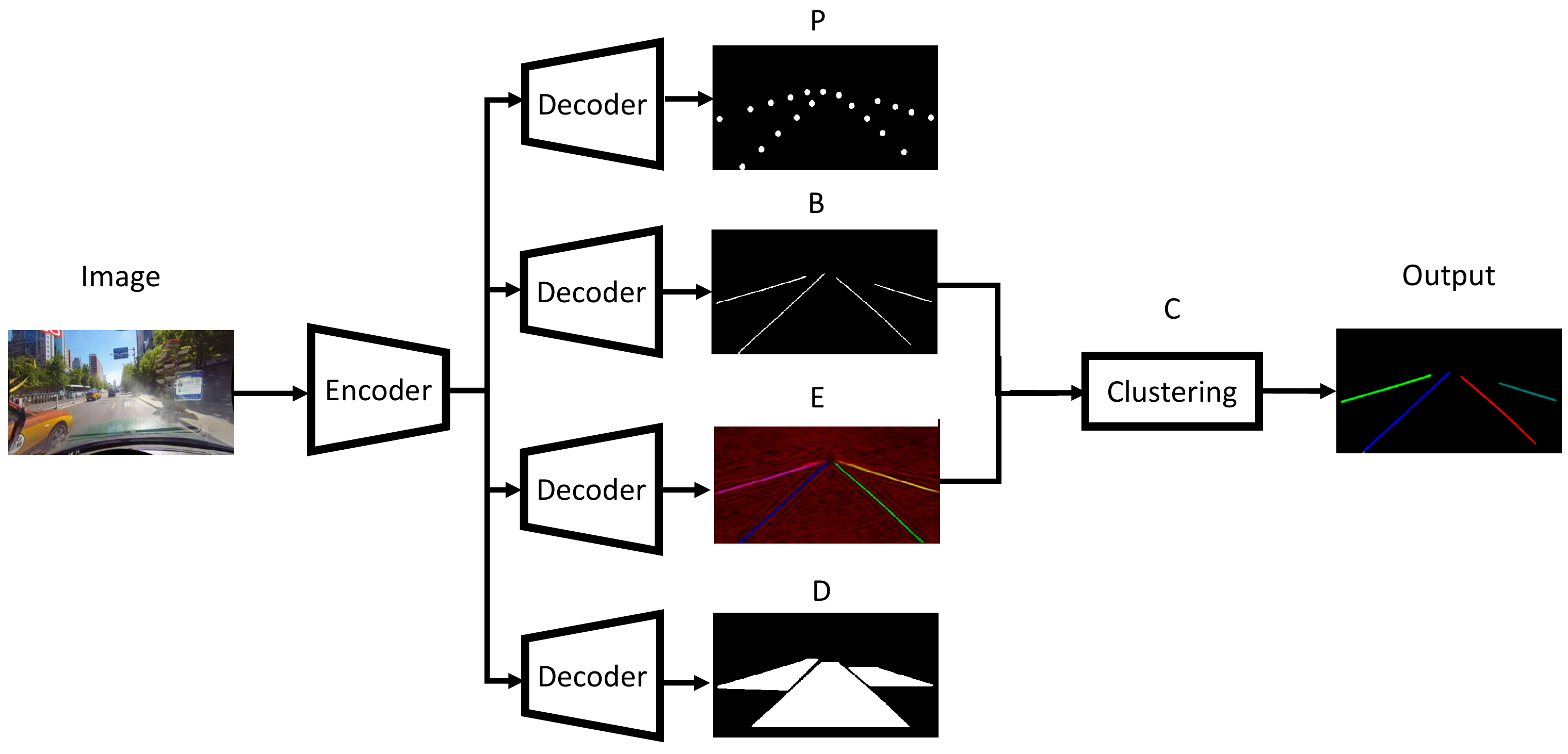}
  \vskip -0.2cm
  \caption{An overview of our agnostic lane detection model.}
  \centering
  \label{fig:pipeline}
\end{figure*}


In this section, we will give a detailed explanation of our framework as shown in Fig.~\ref{fig:pipeline}. Our model is mainly composed of five components, i.e., the binary segmentation branch (B), the drivable area detection branch (D), the lane point regression branch (P), the lane pixel embedding branch (E) and the clustering branch (C). The encoder and decoder of the first four branches are the same but only the encoder is shared.    

\subsection{Binary Segmentation}

\begin{figure}[t]
  \centering
  \includegraphics[width=1.0\linewidth]{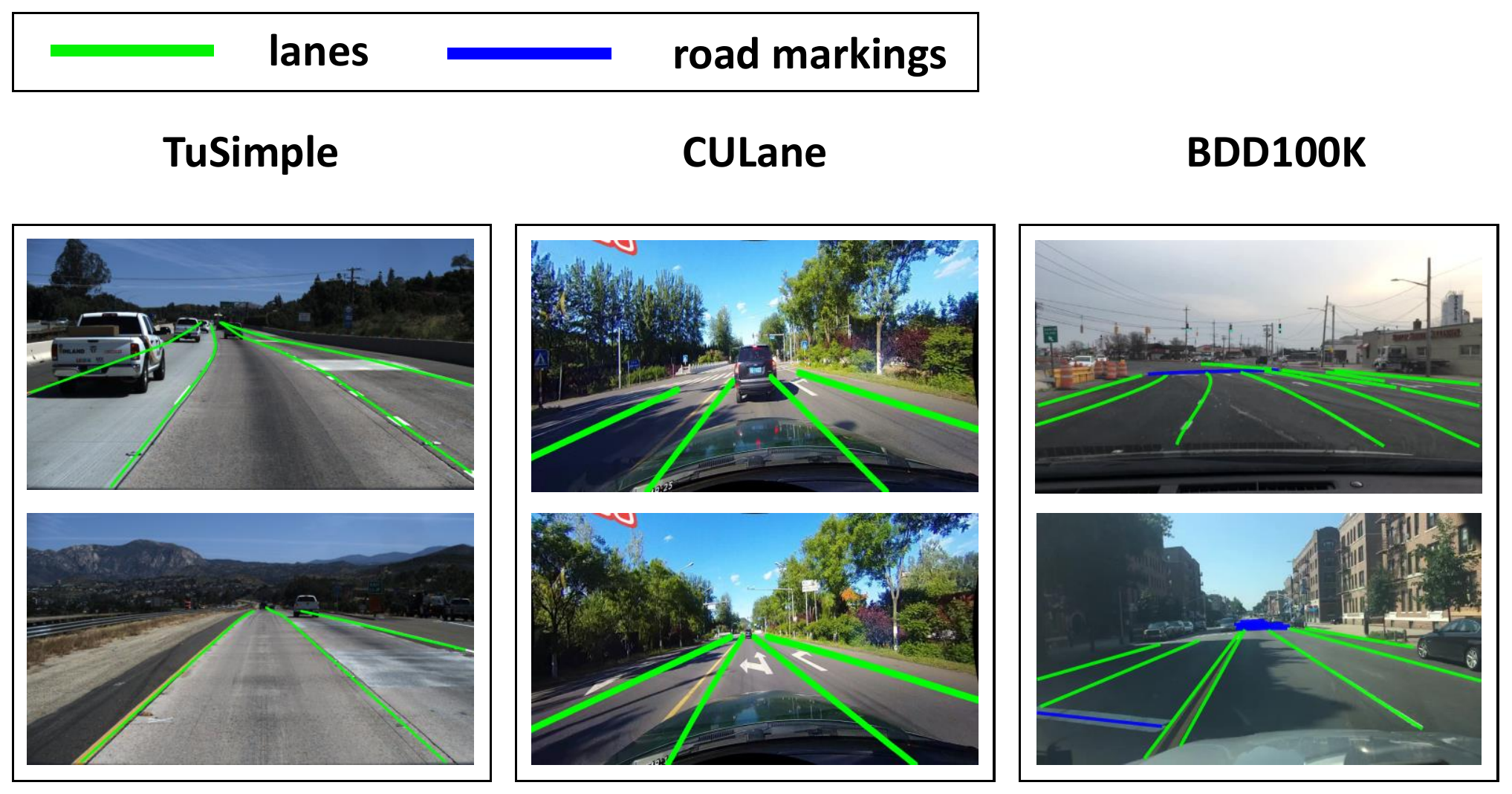}
  \vskip -0.2cm
  \caption{Typical video frames of TuSimple, CULane and BDD100K datasets.}
  \centering
  \label{fig:dataset}
\end{figure}

The objective of the binary segmentation branch is to generate a binary segmentation map, indicating whether each pixel in the original image belongs to the lanes or not. Since the ground-truth lane labels of three datasets are all lane points, we generate the final targets by connecting lane points into lines (see the final targets in Fig.~\ref{fig:dataset}). We use standard cross-entropy loss to train this branch. Besides, to solve the class imbalance of lane pixels and background pixels, the loss of background is multiplied by $0.4$. Moreover, the feature pyramid architecture ~\cite{lin2017feature} is adopted to detect extremely thin lanes.       

\subsection{Drivable Area Detection}

The target of the drivable area detection branch is to output a segmentation map, indicating which part of the road is drivable (we merge the original alternative areas into the drivable areas to provide denser targets). Standard cross-entropy loss is adopted to train this branch. This branch aims at using the boundary of drivable areas to refine the binary segmentation result via providing more structural information.    

\subsection{Lane Point Regression}

The objective of this branch is to regress the position of each lane points. Since the lane points are relatively sparse, we use an 11 x 11 kernel to smooth the original lane point maps to get the final targets of this branch. $L_{2}$ loss is used to train this branch. This branch aims at refining the output of the binary segmentation branch.

\subsection{Lane Pixel Embedding}

The input of this branch is the lane pixels extracted from the binary segmentation maps. We treat each lane in the image as an instance. The target of this branch is to classify the lane pixels into different lane instances. The core idea is that pixels belonging to the same lane instance should be close to each other while those belonging to different lane instances should be far from each other. We utilize the following equation to compute the clustering loss~\cite{neven2018towards}:

\begin{equation}
L_{var}  =  \frac{1}{L} \sum_{c=1}^{L}\frac{1}{N_{c}} \sum_{i=1}^{N_{c}}[\| \mu_{c} - x_{i} \|_{2}^{2} - \delta_{v}]_{+}^{2} ,
\end{equation}

\begin{equation}
L_{dist}  =  \frac{1}{L(L-1)} \sum_{c_{A}=1}^{L} \sum_{c_{B}=1, c_{A} \neq c_{B}}^{L}[\delta_{d} - \| \mu_{c_{A}} - \mu_{c_{B}} \|_{2}^{2}]_{+}^{2} ,
\end{equation}

where $L$ denotes the number of lanes, $x_{i}$ is the embedding of a pixel, $N_{c}$ is the number of elements in cluster $c$, $\mu_{c}$ is the mean embedding of cluster $c$ and $[x]_{+} = max(0, x)$. The first loss term $L_{var}$ is used to keep the distance between pixels belonging to the same lane instance closer than 2$\delta_{v}$. The second loss term $L_{dist}$ is used to keep the distance between different lane clusters farther than $\delta_{d}$. 

\subsection{Clustering}

The clustering branch is used to process the output of the lane pixel embedding branch. In the experiments, we set $\delta_{d} > 6\delta_{v}$. Therefore, given the output of the lane pixel embedding branch, we can randomly select a pixel as the starting point, and then label all pixels whose distance from the selected pixel is smaller than 2$\delta_{v}$ as the same instance. This process is continued until all the lane pixels are assigned to a specific lane instance. Note that this branch does not have any learnable parameters. 

\subsection{Training strategy}

Currently, we adopt a two-stage training strategy. In the first stage, we fix the parameters of branch E and train the branch P, B and D. In the second stage, we fix the parameters of branch P, B and D and train branch E.

\section{Experiments}

In this section, we will first give a brief introduction to three datasets used for evaluation. Then, preliminary experimental results are given.

\subsection{Dataset}

\begin{table*}[!t]
\caption{A brief description about three lane detection datasets.}
\label{dataset_table}
\centering
\begin{tabular}{c|c|c|c|c|c|c|c}
\hline
Name & \# Frame & Train & Validation & Test & Resolution & Road Type & \# Lane $\leq$ 5 ? \\
\hline
TuSimple & 6, 408 & 3, 626 & -- & 2, 782 & 1280 $\times$ 720 & highway & $\surd$ \\
\hline
CULane & 133, 235 & 88, 880 & 9, 675 & 34, 680 & 1640 $\times$ 590 & urban, rural and highway & $\surd$ \\
\hline
BDD100K & 80, 000 & 70, 000 & -- & 10, 000 & 1280 $\times$ 720 & urban, rural and highway & $\times$ \\
\hline
\end{tabular}
\end{table*}

Table \ref{dataset_table} records the basic information of three lane detection datasets. Note that the last column of Table \ref{dataset_table} shows that TuSimple and CULane have no more than 5 lanes in a video frame while BDD100K typically has more than 8 lanes in a video frame. Besides, TuSimple is relatively easy while CULane and BDD100K are more challenging considering the total number of video frames and road types.

\subsection{Evaluation Criterion}

\subsubsection{TuSimple}

In TuSimple dataset, we use the official metric (accuracy) as the evaluation criterion. Besides, false positive ($FP$) and false negative ($FN$) are also listed. The following is the equation to compute accuracy~\cite{tusimple}: 

\begin{equation}
Accuracy = \frac{N_{pred}}{N_{gt}} ,
\end{equation}

where $N_{pred}$ is the number of correctly predicted lane points and $N_{gt}$ is the number of ground-truth lane points.

\subsubsection{CULane and BDD100K}

To judge whether a lane is correctly detected, we treat each lane as a line with fixed pixel width (30 for CULane and 8 for BDD100K) and compute the intersection-over-union (IoU) between labels and predictions. Predictions whose IoUs are larger than 0.5 are considered as true positives (TP). Then, we use $F_{1}-measure$ as the evaluation metric formulated as follows:

\begin{equation}
F_{1}-measure = \frac{2 \times Precision \times Recall}{Precision + Recall} ,
\end{equation}

where $Precision = \frac{TP}{TP + FP}$ and $Recall = \frac{TP}{TP + FN}$.


\subsection{Lane detection model}

We choose ENet~\cite{paszke2016enet} as the backbone model (i.e., the encoder and decoder module in Fig.~\ref{fig:pipeline}). Adam~\cite{kingma2014adam} is selected as the optimizer to train our model with an initial learning rate of $5 \times 10^{-4}$.   

\subsection{Preliminary results on TuSimple and CULane}

\begin{table}[!t]
\caption{Performance of different algorithms on TuSimple testing set.}
\label{tusimple_table}
\centering
\begin{tabular}{c|c|c|c}
\hline
Algorithm & Accuracy & FP & FN \\
\hline \hline
SCNN~\cite{pan2017spatial} & 0.9653 & 0.0617 & 0.0180 \\
\hline
LaneNet~\cite{neven2018towards} & 0.9638 & 0.0780 & 0.0244 \\
\hline
EL-GAN~\cite{ghafoorian2018gan} & 0.9639 & 0.0412 & 0.0336 \\
\hline \hline
\textbf{ENet (ours)} & 0.9629 & 0.0722 & 0.0218 \\
\hline
\end{tabular}
\end{table}

Table \ref{tusimple_table} records the performance of some baselines and our algorithm in the testing set of TuSimple. Since TuSimple is relatively easy and our ENet model has much fewer parameters compared with SCNN (see Table \ref{culane_model}), the performance of our model is satisfying. Table \ref{culane_table} records the performance of some baselines and our algorithms in the testing set of CULane. As can be seen in Table \ref{culane_model}, in terms of the running time efficiency and the number of parameters, our algorithm obviously outperforms other baselines. 

\begin{table*}[!t]
\caption{Performance ($F_{1}$-measure) of different algorithms on CULane testing set. $\dag$ indicates the results are copied from~\cite{pan2017spatial}. For crossroad, only FP is shown.}
\label{culane_table}
\centering
\begin{tabular}{c|c|c|c|c|c|c|c|c|c}
\hline
Category & \textbf{ENet} & ResNet-18 & VGG-16$\dag$ & ReNet$\dag$ & DenseCRF$\dag$ & MRFNet$\dag$ & ResNet-50$\dag$ & ResNet-101$\dag$ & SCNN$\dag$ \\
\hline \hline
Normal & 88.4 & 89.8 & 83.1 & 83.3 & 81.3 & 86.3 & 87.4 & 90.2 & \textbf{90.6} \\
\hline
Crowded & 67.0 & 68.1 & 61.0 & 60.5 & 58.8 & 65.2 & 64.1 & 68.2 & \textbf{69.7} \\
\hline
Night & 61.4 & 64.2 & 56.9 & 56.3 & 54.2 & 61.3 & 60.6 & 65.9 & \textbf{66.1} \\
\hline
No line & 42.9 & 42.5 & 34.0 & 34.5 & 31.9 & 37.2 & 38.1 & 41.7 & \textbf{43.4} \\
\hline
Shadow & 63.4 & 67.5 & 54.7 & 55.0 & 56.3 & 59.3 & 60.7 & 64.6 & \textbf{66.9} \\
\hline
Arrow & 81.9 & 83.9 & 74.0 & 74.1 & 71.2 & 76.9 & 79.0 & 84.0 & \textbf{84.1} \\
\hline
Dazzle light & 57.4 & 59.8 & 49.9 & 48.2 & 46.2 & 53.7 & 54.1 & \textbf{59.8} & 58.5 \\
\hline
Curve & 62.6 & 65.5 & 61.0 & 59.9 & 57.8 & 62.3 & 59.8 & \textbf{65.5} & 64.4 \\
\hline
Crossroad & 2768 & 1995 & 2060 & 2296 & 2253 & \textbf{1837} & 2505 & 2183 & 1990 \\
\hline
Total & 68.8 & 70.5 & 63.2 & 62.9 & 61.0 & 67.0 & 66.7 & 70.8 & \textbf{71.6} \\
\hline
\end{tabular}
\end{table*}

\begin{table}[!t]
\caption{The running time and parameters of different algorithms on CULane testing set.}
\label{culane_model}
\centering
\begin{tabular}{c|c|c|c|c}
\hline
Indicator & \textbf{ENet} & ResNet-18 & ResNet-101 & SCNN \\
\hline \hline
Running time (ms) & \textbf{13.4} & 25.3 & 171.2 & 133.5  \\
\hline
Parameter (M) & \textbf{0.98} & 12.41 & 52.53 & 20.72  \\
\hline
\end{tabular}
\end{table} 

\section{Conclusion}

In this study, we first point out the value and main challenges of the lane detection task. Then, the strengths and weaknesses of both conventional and deep learning based methods are presented. To overcome the shortcomings of previous methods, our agnostic lane detection model is proposed, which utilizes a multi-task learning paradigm and the feature pyramid architecture to exploit structural and contextual information. We use three popular benchmarks, i.e., TuSimple, CULane and BDD100K, to validate the effectiveness of the proposed algorithm. Preliminary experimental results have shown that our model outperforms previous approaches in terms of the running time efficiency and the number of parameters. However, this is still an on-going project and more performance gains will be achieved via a good deployment of different components and a more rational training strategy.     

\section{Declarations and Acknowledgement}

I declare that this report is solely written by me without help from others. In addition, I am the main contributor to the techniques and methodology expounded in this report. The structure of the overall framework, the training strategy as well as the experiments are solely proposed and implemented by me. Besides, I thank Professor Chen Change Loy, Dr. Chunxiao Liu and Dr. Zheng Ma for their helpful discussions during the whole project.

\bibliographystyle{plain}
\bibliography{first_year_thesis_hyn}
\end{document}